\documentclass[sigconf]{acmart}
\AtBeginDocument{%
  }

\setcopyright{acmlicensed}
\copyrightyear{2025}  
\acmYear{2025}  
\setcopyright{cc}
\setcctype{by-nc}
\acmConference[CIKM '25]{Proceedings of the 34th ACM International Conference on Information and Knowledge Management}{November 10--14, 2025}{Seoul, Republic of Korea}
\acmBooktitle{Proceedings of the 34th ACM International Conference on Information and Knowledge Management (CIKM '25), November 10--14, 2025, Seoul, Republic of Korea}\acmDOI{10.1145/3746252.3761304}
\acmISBN{979-8-4007-2040-6/2025/11}

\usepackage{amsmath,amsfonts}
\usepackage{algorithm}
\usepackage{hyperref} 
\usepackage{algorithmic}
\usepackage{multirow}
\usepackage{pifont}
\usepackage{enumitem}
\usepackage{subfigure}
\usepackage{array}
\usepackage{textcomp}
\usepackage{stfloats}
\usepackage{url}
\usepackage{verbatim}
\usepackage{graphicx}
\usepackage{float}
\usepackage{colortbl}
\usepackage{xcolor}
\definecolor{lightcyan}{rgb}{0.88, 1.0, 1.0}

\begin{document}

\title{OASIS: Open-world Adaptive Self-supervised and Imbalanced-aware System}

\author{Miru Kim}
\thanks{M. Kim and M. Joe contributed equally. Corresponding author: M. Kwon. All authors are with the Department of Intelligent Semiconductors, and M. Kwon is also affiliated with the School of Electronic Engineering.}
\affiliation{%
  \institution{Soongsil University}
  \streetaddress{Address}
  \city{Seoul}
  \country{South Korea}}
\email{mirukim00@soongsil.ac.kr}

\author{Mugon Joe}
\affiliation{%
  \institution{Soongsil University}
  \streetaddress{Address}
  \city{Seoul}
  \country{South Korea}}
\email{mugon@soongsil.ac.kr}

\author{Minhae Kwon}
\affiliation{%
  \institution{Soongsil University}
  \streetaddress{Address}
  \city{Seoul}
  \country{South Korea}}
\email{minhae@ssu.ac.kr}


\begin{abstract}
The expansion of machine learning into dynamic environments presents challenges in handling open-world problems where label shift, covariate shift, and unknown classes emerge concurrently. Post-training methods have been explored to address these challenges, adapting models to newly emerging data. However, these methods struggle when the initial pre-training is performed on class-imbalanced datasets, limiting generalization to minority classes. To address this, we propose \textbf{OASIS}, an \textbf{O}pen-world \textbf{A}daptive \textbf{S}elf-supervised and \textbf{I}mbalanced-aware \textbf{S}ystem. OASIS consists of two learning phases: pre-training and post-training. The pre-training phase aims to improve the classification performance of samples near class boundaries via a novel borderline sample refinement step. Notably, the borderline sample refinement step critically improves the robustness of the decision boundary in the representation space. Through this robustness of the pre-trained model, OASIS generates reliable pseudo-labels, adapting the model against open-world problems in the post-training phase. Extensive experiments demonstrate that our method significantly outperforms state-of-the-art post-training techniques in both accuracy and efficiency across diverse open-world scenarios.
\end{abstract}

\ccsdesc[500]{Computing methodologies~Artificial intelligence}
\ccsdesc[500]{Computing methodologies~Semi-supervised learning settings}
\ccsdesc[500]{Computing methodologies~Online learning settings}

\keywords{Open-world problem, Borderline sample refinement, Semi-supervised learning, Online post-training}


\maketitle

\section{Introduction}
\begin{figure}[t]
    \centering    
    \includegraphics[width=1\columnwidth]{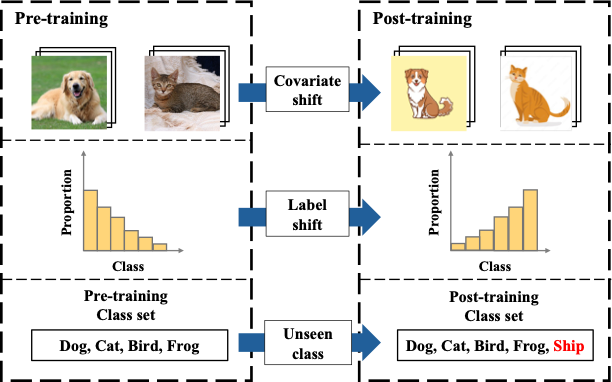}
    \caption{Illustration of open-world challenges in machine learning}
    \label{fig:openworld}
\end{figure}

The rapid advancement of machine learning technologies has propelled their application into environments where data distributions shift over time and unseen classes emerge, challenging traditional models that assume stationary data and closed label sets. This transition into open-world scenarios exposes several critical limitations in existing methodologies, motivating research on open-world adaptation~\cite{UDA,UNIDA,ATLAS,OSLS,OWSSL,open1,open2,open3,open4,open5}. The open-world challenges are illustrated in Figure~\ref{fig:openworld}, and a comparison of the open world problem settings is summarized in Table~\ref{tab:param_reduction}.

\begin{table}[t]
    \centering
    \caption{Comparison of different problem settings.}
    \begin{tabular}{l|cccc}
        \hline
         & Covariate & Label  & Seen & Class  \\
            Setting & Shift & Shift & Detection &  Imbalance \\
        \hline
        UDA~\cite{UDA} & \checkmark & - & - & -\\
        ATLAS~\cite{ATLAS} & - & \checkmark & - & -\\     
        UNIDA~\cite{UNIDA} & \checkmark & - & \checkmark & -\\
        OSLS~\cite{OSLS} & - & \checkmark & \checkmark & -\\
        OW-SSL~\cite{OWSSL} & \checkmark & \checkmark & \checkmark & -\\\hline
        \textbf{Proposed} & \checkmark & \checkmark & \checkmark & \checkmark\\        
        \hline
    \end{tabular}
    \label{tab:param_reduction}
\end{table}

Early efforts in open-world adaptation primarily addressed individual distribution shifts in isolation. For instance, some approaches focused on covariate shift, where distributions of input features between the training and inference differ across domains due to variations in environmental factors, data collection conditions, or domain differences~\cite{cov1,cov2,UDA,CIKM1,CIKM2,CIKM3}. Others tackled label shift, which occurs when the prior distribution of classes in the training data differs from that in the test data~\cite{OLS1,OLS2,ATLAS,CIKM4,CIKM5,CIKM6}. This shift is particularly problematic in open-world settings, where some classes may become more prevalent while others become rare or even absent in new environments. Addressing these challenges requires models to continuously adjust to shifting distributions, ensuring that they remain aligned with the changing data environment.

The presence of unknown classes further complicates open-world learning, as models can encounter new classes during post-deployment~\cite{UNIDA,OSLS,unseen1,unseen2}. To address unseen class detection, it is effective to classify inputs as unknown when they deviate significantly from the patterns learned during pre-training. Open-world learning must incorporate strategies for reliable unseen class detection and pseudo-labeling to continuously expand the model’s knowledge base while maintaining robustness~\cite{unseen3,unseen4,OWSSL}. However, this robustness is difficult to achieve when the pre-training data suffers from class imbalance, as it hampers the model’s ability to form robust decision boundaries, especially near class borders.

In response to these challenges, we propose a novel framework, OASIS, which enhances representation learning through a borderline sample refinement step to establish a solid foundation under class-imbalance data and adapts the model via reliable pseudo-labeling derived from this foundation. Our method begins with a contrastive-based pre-training phase that improves the representation of borderline samples. This phase encourages borderline samples to move closer to their corresponding class centers, reducing intra-class variance in the representation space. After pre-training, OASIS performs self-supervised post-training using pseudo-labeling to effectively handle the complexities of open-world adaptation without requiring extensive manual annotations. With a refined representation space established during pre-training, the model is able to generate reliable pseudo-labels for newly emerging data. This self-supervised mechanism enhances the model’s ability to incorporate unseen classes and mitigate distribution shifts, including label and covariate shifts.

The main contributions of this paper are as follows.
\begin{itemize}
    \item We propose OASIS, which effectively addresses open-world problems involving label shift, covariate shift, and unseen classes.
    \item We propose a simple, but powerful, borderline refinement step in the pre-training phase. This step improves the decision boundary of each class in the representation space, which leads to a strong foundation for more effective adaptation during post-training.
    \item We propose an imbalance-aware pre-training method based on contrastive learning that enhances the representation of minority classes. 
    \item We propose a pseudo-labeling method based on a refined decision boundary in the post-training phase. This self-supervised approach allows the model to adapt to open-world problems without requiring labeled data during post-training. 
    \item We propose a conditional adaptation for the post-training, which reduces computational burden. 
    \item Extensive simulations demonstrate that our framework significantly outperforms existing post-training methods in terms of accuracy and efficiency across various open-world scenarios.
\end{itemize}
Commonly used notations are summarized in Appendix~\ref{sec:simul_setup}.

\section{Related Works}
\hspace{1em} \textbf{Seen Detection.}
In open-world scenarios, identifying whether a sample belongs to a known or unknown category is a crucial challenge~\cite{openworld,OSR,OWSSL,prototype1,prototype2}. Open-world learning methods ~\cite{openworld} address this issue by designing classifiers that reject unknown samples while correctly classifying known ones. A related concept, open set recognition (OSR) ~\cite{OSR}, focuses on detecting out-of-distribution samples without explicitly clustering novel categories. More recently, open-world semi-supervised learning (OW-SSL) ~\cite{OWSSL} has been proposed, where classification and novel class discovery are tackled simultaneously. These methods often employ prototype-based learning ~\cite{prototype1,prototype2} to refine their decision boundaries. Our approach extends these methodologies by leveraging a contrastive pre-training phase to improve the seen class classification while integrating a pseudo-labeling strategy to detect the unknown class.

\textbf{Label Shift. }
Changing label distributions over time poses a major challenge, as post-training distributions often differ from training data, leading to reduced model performance~\cite{ODS,FTH,ATLAS}. Various methods have been proposed to address this, including the follow the history (FTH) algorithm, which averages past label distributions for gradual shifts, and the follow the fixed window history (FTFWH) approach, which adapts quickly to recent data but is sensitive to window size~\cite{FTH}. The unbiased online gradient descent (UOGD) algorithm builds on online gradient descent (OGD) with an unbiased risk estimator, enabling continuous updates without labeled data but struggling with rapid shifts due to its fixed learning rate~\cite{ATLAS}. The adapting to label shift (ATLAS) algorithm improves adaptability through an ensemble approach, dynamically combining base learners to optimize performance under shifting distributions~\cite{ATLAS}. Additionally, pseudo-labeling methods have been employed to address data scarcity, leveraging model predictions as training labels.

\textbf{Covariate Shift.}
Covariate shift occurs when the feature distributions of training and test datasets differ, often leading to significant degradation in model performance~\cite{CIKM7,CIKM8,Neurips1,Neurips2,Neurips3}. Traditional unsupervised domain adaptation (UDA) methods aim to mitigate this issue by learning domain-invariant representations ~\cite{UDA}. One common approach is adversarial domain adaptation, where a discriminator is trained to align the feature distributions of source and target domains ~\cite{adversarial1, adversarial2}. However, these methods typically assume a fixed set of categories, making them ineffective in open-world settings where novel classes may emerge. Universal domain adaptation (UniDA) ~\cite{UNIDA} extends UDA by allowing the model to handle both known and unknown categories in the target domain. Despite its advancements, UniDA still struggles with mechanisms to discover novel classes effectively.

\textbf{Class Imbalance.}
Real-world datasets frequently exhibit long-tailed distributions, where some classes have significantly fewer samples than others~\cite{Neurips4,Neurips5,imbalance2,Cont1,Cont2}. Traditional approaches for class imbalance involve re-sampling techniques ~\cite{imbalance1} or cost-sensitive learning ~\cite{imbalance2}. However, these methods are often ineffective in open-world settings, where the distribution shift between training and test data further exacerbates imbalance issues. Contrastive learning has emerged as a powerful technique to mitigate class imbalance by enforcing more structured feature representations ~\cite{Cont1,Cont2}. 

\section{Problem Formulation}
In this section, we provide a formal description of the data for the pre-training and post-training phases, including novel class, label shift, and covariate sift. Next, we provide the learning objectives of the proposed solution. The pre-training is conducted in timestep $t=0$, and post-training is processed from $t=1$ to $t=T$.

\subsection{Open-world Scenario} 
We consider the labeled dataset $\mathcal{D}_0$ for pre-training and unlabeled dataset $\mathcal{D}_t$ for post-training at timestep $t~(1 \leq t \leq T)$. The labeled dataset $\mathcal{D}_0 = \{\mathbf{x}_i^0, y_i^0\}_{i=1}^{N_0}$ consist of data $x_i^0$ and label $y_i^0$ with number of data samples $N_0$. In labeled dataset $\mathcal{D}_0$, the label $y_i^0$ is come from known class $\mathcal{C}_{0} \subset \mathcal{C}_{all}$, where $\mathcal{C}_{all}$ represents the all classes. The labeled dataset $\mathcal{D}_0$ follows the class-imbalance label distribution $\omega_0$.

The unlabeled dataset $\mathcal{D}_t = \{x_i^t\}_{i=1}^{N_t}$ consist of only data $x_i^t$ without label. Here, $N_t$ denotes the number of data samples at timestep $t$. The label of each sample $x_i^t$ in unlabeled dataset $\mathcal{D}_t$ can be a known class from $\mathcal{C}_{0}$ or a novel class from $\mathcal{C}_{all} \backslash \mathcal{C}_{0}$. Specifically, we can model the data shift of unlabeled datasets, including novel classes, over timestep $t~(1\leq t \leq T)$ in the open-world as follows.
\begin{equation}
    \Omega^t(c) = \alpha^t \omega_0(c) + (1-\alpha^t) \omega_T(c)
\end{equation}
Here, $c$ represents the possible class from all classes $\mathcal{C}_{all}$. $\omega_0(c)$ and $\omega_T(c)$ represent the label distribution of the pre-training dataset and final dataset at timestep $T$, respectively, and $\alpha^t$ controls the data shift.\footnote{We provide four types of $\alpha^t$ setting in the simulation section.} $\omega_0(c)=0$ holds for the novel class $c \notin \mathcal{C}_0$, indicating its absence in the labeled dataset. 


\subsection{Learning Objectives}
The proposed framework involves training a model $\theta$, with two objectives: pre-training and post-training. The objective of the pre-training phase is to generate a model that contains robust performance across known class $\mathcal{C}_0$ in a labeled dataset $\mathcal{D}_0$ following the class-imbalanced label distribution $\omega_0$. To achieve this objective, we propose the class-imbalance aware contrastive learning to update the model $\theta$.


The objective of the post-training phase is to adapt the model for the unlabeled dataset $\mathcal{D}_t$ at each timestep $t$, which includes managing covariate shift, label shift, and detecting novel classes. To achieve the objective of post-training, we divide the model $\theta$ into frozen parameters $f$ and learnable parameters $l$, and propose a learning method based on self-supervised training for the learnable parameter $l$.


\begin{figure*}[t]
    \centering
    \begin{minipage}{0.48\textwidth}
        \centering
        \includegraphics[width=\textwidth]{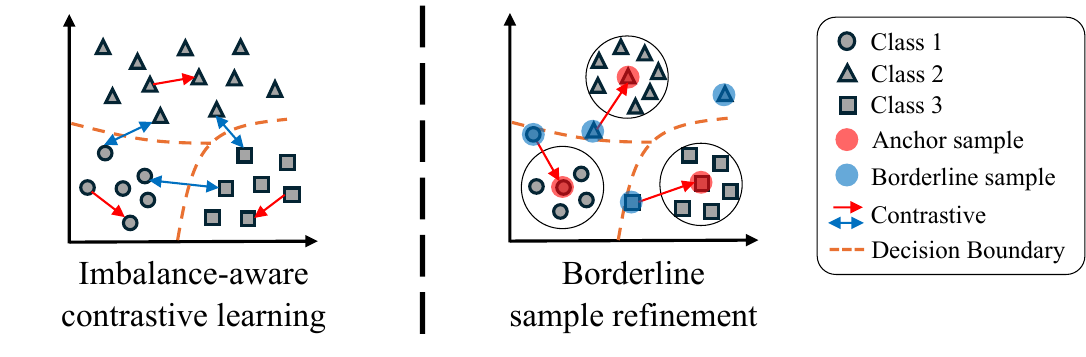}
        \vfill
        \scriptsize{(a) Pre-training Phase}
    \end{minipage}
    \hfill
    \begin{minipage}{0.48\textwidth}
        \centering
        \includegraphics[width=\textwidth]{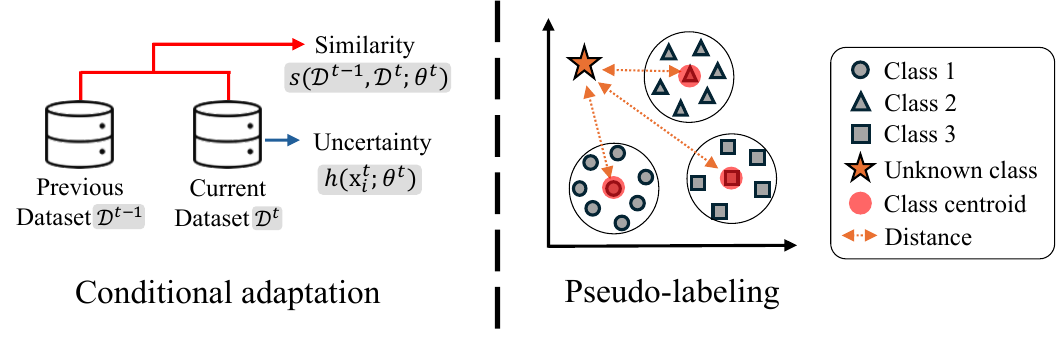}
        \vfill
        \scriptsize{(b) Post-training Phase}
    \end{minipage}
    \hfill
    \caption{Overview of the proposed solutions: (a) Pre-training phase (b) Post-training phase}
    \label{fig:fig2}
\end{figure*}


\section{Imbalance-aware Pre-training with Borderline Refinement}
The pre-training phase is structured to develop a model capable of handling class imbalance while refining borderline samples. It consists of two primary steps: imbalance-aware training and borderline sample refinement. In the imbalance-aware training stage, both the frozen parameter $f$ and the learnable parameter $l$ are optimized to ensure robust performance across the inherently imbalanced classes in dataset $\mathcal{D}_0$. This process enhances the model’s sensitivity to minority classes while maintaining overall accuracy. Subsequently, the proposed borderline sample refinement step improves representation learning by concentrating on samples near class boundaries. This step enhances the decision boundary for each class within the representation space. By acquiring improved representations, the model creates a solid foundation for more efficient adaptation in post-training. The total progress of the proposed pre-training phase is described in Algorithm~\ref{alg:pretraining}.

\subsection{Step I: Imbalance-aware Contrastive Learning}
In this step, the objective is to mitigate the class imbalance issue through a contrastive learning framework. Within this framework, the selection of data index pairs $(i, j)$ is performed iteratively $N_0$ times. The first index $i$ is chosen sequentially, ensuring that $\forall (\mathbf{x}_i^0, {y}_i^0) \in \mathcal{D}_0$, the selection follows the label distribution $\Omega^0(c)$ of the pre-training phase such that $y_i^0 \sim \Omega^0(c)$. The second sample $(\mathbf{x}_j^0, {y}_j^0)$ is selected based on the distribution of $1-\Omega^0(c)$, i.e., $y_j^0 \sim 1-\Omega^0(c)$, assigning a higher selection probability to minority classes and a lower probability to majority classes. This approach ensures a more balanced pairing of $i$ and $j$ between the majority and minority classes.

In the first step, the pre-training loss $\mathcal{L}_{\text{pre}}(\cdot)$ is defined as follows. 
\begin{align} 
\label{equ1}
\mathcal{L}_{\text{pre}}(\mathbf{x}_i^0, \mathbf{x}_j^0, &y_i^0, y_j^0; \theta^0) \notag \\
  = &\lambda \Big( \mathcal{L}_{\text{class}} \left(\mathbf{x}_i^0, y_i^0; \theta^0\right) + \mathcal{L}_{\text{class}} \left(\mathbf{x}_j^0, y_j^0; \theta^0\right) \Big) \notag \\
 &  + (1 - \lambda) \mathcal{L}_{\text{rep}} \left(\mathbf{x}_i^0, \mathbf{x}_j^0, y_i^0, y_j^0; \theta_{:\bar L}^0\right)
\end{align}
It is a combination of the cross-entropy loss $\mathcal{L}_{\text{class}}(\cdot)$ for optimizing classification performance and contrastive loss $\mathcal{L}_{\text{rep}}(\cdot)$ for learning representation. Here, \( \lambda \) for ($0\le \lambda \le 1$) is a scaling parameter that balances the two objectives. 

The cross-entropy loss \( \mathcal{L}_{\text{class}}(\mathbf{x}_i^0, y_i^0; \theta^0) \) is defined as follows.
\begin{equation} 
\label{equ2}
\mathcal{L}_{\text{class}} (\mathbf{x}_i^0, y_i^0; \theta^0) = - \sum_{c \in \mathcal{C}_0} p_{\mathbf x_i^0}(c) \log q_{\mathbf{x}_i^0}(c; \theta^0)
\end{equation}
The objective of \eqref{equ2} is to minimize the gap between the probability of the true label \( p_{\mathbf x_i^0}(c) \) and the model output \( q_{\mathbf{x}_i^0}(c; \theta^0) \). 

Next, the contrastive loss \( \mathcal{L}_{\text{rep}}(\mathbf{x}_i^0, \mathbf{x}_j^0, y_i^0, y_j^0; \theta_{:\bar L}^0) \) in \eqref{equ1}
is computed in the latent space, which is the output of the \( \bar{L} \)-th layer for $1 < \bar L < L$.\footnote{The contrastive representation layer $\bar L$ is chosen at the middle layer of the model, which is the end of the feature extraction and the beginning of classification.}  
\begin{align}   
&\mathcal{L}_{\text{rep}}(\mathbf{x}_i^0, \mathbf{x}_j^0, y_i^0, y_j^0; \theta_{:\bar L}^0) \label{eqn:l_cont} \\ 
&= \Big[ \mathbb{I}_{\{y_i^0 = y_j^0\}} \Big(\|\theta_{\bar{:L}}^0(\mathbf{x}_i^0) - \theta_{\bar{:L}}^0(\mathbf{x}_j^0)\|_2 \Big)   \label{eqn:same_pair} \\
& +\mathbb{I}_{\{y_i^0 \neq y_j^0\}}  \Big(\max\left(0, \epsilon - \|\theta_{\bar{:L}}^0(\mathbf{x}_i^0) - \theta_{\bar{:L}}^0(\mathbf{x}_j^0)\|_2\right)\Big) \Big]\label{equ3}
\end{align}
Here, \eqref{eqn:same_pair} encourages the model to reduce the distance between samples of the same class, while \eqref{equ3} ensures a minimum separation of \(\epsilon\) between samples of different classes. The Euclidean distance \( \|\cdot\|_2 \) is used to measure distances in the latent space, with the margin hyperparameter \( \epsilon \geq 0 \) controlling the separation between different class samples. The indicator function \( \mathbb{I}_{\{condition\}} \) is defined as follows.
\begin{align*}
\mathbb{I}_{\{condition\}} = 
\begin{cases} 
1, & \text{if the $condition$ is true} \\ 
0, & \text{otherwise}
\end{cases}
\end{align*}

The model \( \theta^0 \) is updated in a supervised manner using the proposed pre-training loss $\mathcal{L}_{\text{pre}}(\mathbf{x}_i^0, \mathbf{x}_j^0, y_i^0, y_j^0; \theta^0)$ in \eqref{equ1}. 
Following this stage, the model is stabilized through the novel borderline sample refinement step.

\begin{algorithm}[t]
\caption{Pre-training phase}
\label{alg:pretraining}
\begin{algorithmic}[1]
    \REQUIRE Dataset $\mathcal{D}_0 = \{\mathbf{x}_i^0, y_i^0\}_{i=1}^{N_0}$, mean vectors $\{ \mu_c\}_{c\in\mathcal C}$, covariance matrices $\{\Sigma_c\}_{c \in \mathcal C}$, borderline threshold \( \phi_{\text{border}} \)
    \ENSURE Trained model \( \theta^0 \)
    
        \STATE \textbf{// Step I: Imbalance-aware Training}
        \FOR{\( i = 1 \) to \( N \)}
        \STATE Select the first sample $(\mathbf{x}_i^0, y_i^0)$
        \STATE Sample the second sample $(\mathbf{x}_j^0, y_j^0) \sim 1-\Omega^0$ \STATE Compute $\mathcal{L}_{\text{pre}}(\mathbf{x}_i^0, \mathbf{x}_j^0, y_i^0, y_j^0; \theta^0)$ in \eqref{equ1}
            \STATE Update model $\theta^0 \leftarrow \theta^0 - \eta \nabla \mathcal{L}_{\text{pre}}(\mathbf{x}_i^0, \mathbf{x}_j^0, y_i^0, y_j^0; \theta^0)$
        \ENDFOR
        
        \STATE \textbf{// Step II: Borderline Sample Refinement}
        \FOR{\( i = 1 \) to \( N \)}
            \STATE Compute $D_{\text{MD}}(\mathbf{x}_i^0, y_i^0, \mu_c, \Sigma_c; \theta_{\bar{:L}}^0)$ in~\eqref{eq:MD-diff}
            
            \IF{\(D_{\text{MD}}(\mathbf{x}_i^0, y_i^0, \mu_c, \Sigma_c; \theta_{\bar{:L}}^0) > \phi_{\text{border}} \)}
                \STATE Assign \( (\mathbf{x}_i^0, y_i^0) \) as the borderline sample \( (\mathbf{x}_{\text{b},c}^0, y_{\text{b},c}^0) \)
              
            \ENDIF
        \ENDFOR
        \FOR{$c \in \mathcal C$}
        \STATE Select anchor sample  $(\mathbf{x}_{\text{a},c}^0, y_{\text{a},c}^0)$ based on \eqref{eqn:anchor}

        \FOR{all borderline samples in class $c$}
        \STATE Compute $\mathcal{L}_{\text{pre}}(\mathbf{x}_{\text{a},c}^0, \mathbf{x}_{\text{b},c}^0, {y}_{\text{a},c}^0, {y}_{\text{b},c}^0; \theta^0)$ in \eqref{equ1}
            \STATE Update model\\ $\theta^0 \leftarrow \theta^0 - \eta \nabla \mathcal{L}_{\text{pre}}(\mathbf{x}_{\text{a},c}^0, \mathbf{x}_{\text{b},c}^0, {y}_{\text{a},c}^0, {y}_{\text{b},c}^0; \theta^0)$
            \ENDFOR
            \ENDFOR
\end{algorithmic}
\end{algorithm}

\subsection{Step II: Borderline Sample Refinement}
In this step, we concentrate on the novel approach to refine the feature representations of borderline samples that blur the decision boundary, guiding them toward their respective class centroids in the latent space. To assess how close a data sample is to its class centroid, we utilize the Mahalanobis distance (MD) at the \(\bar{L}\)-th layer~\cite{kye2022hierarchical, latentMD}. When the degree of class imbalance increases, MD plays an important role in standardizing the distance by ensuring unit variance for each class, making it particularly effective for handling imbalanced data with varying sample sizes and dispersion across classes.

Let $D_{\text{MD}}(\mathbf{x}_i^0, y_i^0, \mu_c, \Sigma_c; \theta_{\bar{:L}}^0)$ denote the MD of $c$ class sample $(\mathbf{x}_i^0, y_i^0)$  at the output of $\bar L$-th layer. It measures a normalized distance from the centroid, defined as follows. 
\begin{align}
D_{\text{MD}}(\mathbf{x}_i^0, y_i^0, &\mu_c, \Sigma_c; \theta_{\bar{:L}}^0) \notag \\ 
&= \sqrt{ \left(\theta_{\bar{:L}}^0(\mathbf{x}_i^0) - \mu_c\right)^T \Sigma_c^{-1} \left(\theta_{\bar{:L}}^0(\mathbf{x}_i^0) - \mu_c\right) } \label{eq:MD-diff}
\end{align}
Here, \( \mu_c = \mathbb{E}\Big[\theta_{\bar{:L}}^0(\mathbf{x}_i^0) \mid c=y_i^0\Big] \) denotes the mean vector of \( c \) class samples, and $\Sigma_c = \mathbb{E}\Big[\left(\theta_{\bar{:L}}^0(\mathbf{x}_i^0) - \mu_c\right )
\Big(\theta_{\bar{:L}}^0(\mathbf{x}_i^0) - \mu_c\Big)^T \Big|  c=y_i^0 \Big]$ denotes the covariance matrix of class \( c \).

\noindent \textbf{Anchor sample selection:} For each class $c\in \mathcal C$, we determine an anchor sample $(\mathbf{x}_{\text{a},c}^0, y_{\text{a},c}^0)$. The anchor sample \( (\mathbf{x}_{\text{a},c}^0, y_{\text{a},c}^0) \) is selected as the sample with the smallest MD within each class \( c \), i.e.,
\begin{align}
\mathbf{x}_{\text{a},c}^0 = \arg\min_{\mathbf{x}_i^0 \in \mathbf{X}^0} D_{\text{MD}}(\mathbf{x}_i^0, y_i^0, \mu_c, \Sigma_c; \theta_{:\bar L}^0).
\label{eqn:anchor}
\end{align}
Because there is one anchor per class, there are $|\mathcal C|$ anchor samples.

\noindent \textbf{Borderline sample selection:} We determine the borderline sample $(\mathbf{x}_{\text{b},c}^0, y_{\text{b},c}^0)$ that require refinement if $D_{\text{MD}}(\mathbf{x}_{\text{b},c}^0, y_{\text{b},c}^0, \mu_c, \Sigma_c; \theta_{:\bar L}^0)$ exceeds a borderline threshold \( \phi_{\text{border}} \), i.e.,
\begin{equation}
D_{\text{MD}}(\mathbf{x}_{\text{b},c}^0, y_{\text{b},c}^0, \mu_c, \Sigma_c; \theta_{:\bar L}^0) > \phi_{\text{border}}.
\end{equation}
The objective of this step is to pull borderline samples closer to their corresponding anchor, thereby sharpening class-specific decision boundaries in the representation space.

\noindent \textbf{Distance Adjustment:} In the borderline sample refinement step, the pre-training loss $\mathcal{L}_{\text{pre}}(\cdot)$ is the same as defined in \eqref{equ1}, but the pair of samples $(\mathbf{x}_{\text{a},c}^0, \mathbf{x}_{\text{b},c}^0, {y}_{\text{a},c}^0, {y}_{\text{b},c}^0)$ is determined as an anchor and a borderline sample for the same class. 
The contrastive loss $\mathcal{L}_{\text{rep}}(\mathbf{x}_{\text{a},c}^0, \mathbf{x}_{\text{b},c}^0, y_{\text{a},c}^0, y_{\text{b},c}^0; \theta_{:\bar{L}}^0)$ in \eqref{eqn:l_cont} can be simplified as follows, since the pair of input sample have the same class.
\begin{align}   \label{equ5}
\mathcal{L}_{\text{rep}}\left(\mathbf{x}_{\text{a},c}^0, \mathbf{x}_{\text{b},c}^0, y_{\text{a},c}^0, y_{\text{b},c}^0; \theta_{:\bar L}^0\right)  = \|\theta_{\bar{:L}}^0(\mathbf{x}_{\text{a},c}^0)  - \theta_{\bar{:L}}^0(\mathbf{x}_{\text{b},c}^0)\|_2
\end{align}
In~\eqref{equ5}, the contrastive loss minimizes the distance between the anchor and borderline samples. This allows refinement of the representation of borderline samples, which helps enhance classification performance. 

The model \( \theta^0 \) is updated in a supervised manner based on the proposed pre-training loss $\mathcal{L}_{\text{pre}}(\mathbf{x}_{\text{a},c}^0, \mathbf{x}_{\text{b},c}^0, y_{\text{a},c}^0, y_{\text{b},c}^0; \theta^0)$ in \eqref{equ1}. Once this step is completed, the model returns to the imbalance-aware contrastive learning step. By alternating these two steps, the model progressively improves its ability to classify both majority and minority classes.

\section{Self-supervised Post-training in Open-world}
Once the model is fully trained, it is deployed in the real world. In this environment, the data label distribution often differs from that of the pre-training dataset, with a higher occurrence of minority class samples. Furthermore, the model faces data samples from an unseen class. These open-world settings can lead to degraded classification accuracy, making post-training necessary. To tackle these issues, we propose a pseudo-labeling and conditional update method. The proposed post-training phase is shown in Fig.~\ref{fig:fig2}, and the procedure is presented in Algorithm~\ref{alg:adaptive_update}.

\begin{algorithm}[t]
\caption{Post-training phase}
\label{alg:adaptive_update}
\begin{algorithmic}[1]
    \REQUIRE Model $\theta^t$, 
    Dataset $\mathcal{D}_t = \{\mathbf{x}_i^0\}_{i=1}^{N_t}$, $\mathcal{D}_{t-1} = \{\mathbf{x}_i^0\}_{i=1}^{N_{t-1}}$, thresholds \( \phi_{\text{ent}} \), \( \phi_{\text{cos}} \)
    \ENSURE Updated model \( \theta^{t+1} \)

    \STATE Compute similarity \( s(\mathcal{D}_t, \mathcal{D}_{t-1}; \theta^t) \)

        \FOR{\( i = 1 \) to \( N^t \)}
        \STATE Compute entropy \( h(\mathbf{x}_i^t; \theta^t) \)
            \IF{ \( h(\mathbf{x}_i^t; \theta^t) \geq \phi_{\text{ent}} \) \textbf{and}
    \( s(\mathcal{D}_t, \mathcal{D}_{t-1}; \theta^t) < \phi_{\text{cos}} \) 
    }
            \STATE Generate pseudo-label \( \tilde{y}_i^t \) based on \eqref{eqn:pseudo-label} 
            \IF{\( \tilde{y}_i^t \) is generated}
                    \STATE Compute $\mathcal{L}_{\text{post}}(\mathbf{x}_i^t, \tilde{y}_i^t; \theta^t)$ 
                    \STATE Update learnable parameters based on~\eqref{eq:post} \\ 
            \ENDIF
        \ENDIF
        \ENDFOR
\end{algorithmic}
\end{algorithm}

\subsection{Conditional Adaptation}
To minimize computational burden after deployment, our proposed framework activates post-training only if the uncertainty level of the model prediction exceeds \( \phi_{\text{ent}} \) and the similarity between the predicted label distribution of $\mathcal{D}^{t}$ and $\mathcal{D}^{t-1}$ is smaller than \( \phi_{\text{cos}} \). 

We first define the uncertainty level as the entropy $h(\mathbf{x}_i^t; \theta^t)$, which quantifies the uncertainty in the model predictions.
\begin{align*}
h(\mathbf{x}_i^t; \theta^t) = - \sum_{c \in \mathcal{C}_0} q_{\mathbf{x}_i^t}(c; \theta^t) \log q_{\mathbf{x}_i^t}(c; \theta^t)
\end{align*}
If $h(\mathbf{x}_i^t; \theta^t)$ exceeds the threshold $\phi_{\text{ent}}$, indicating a higher degree of uncertainty, the model must be fine-tuned to improve the prediction confidence.

Next, we measure the similarity of predicted label distribution between $\mathcal{D}^{t}$ and $\mathcal{D}^{t-1}$.  
If the similarity is below the threshold \( \phi_{\text{cos}} \), this suggests potential shifts in the label distribution, making post-training necessary.
We denote the similarity $s(\mathcal{D}^t, \mathcal{D}^{t-1}; \theta^t)$ as the cosine similarity between the outputs of the current data $\mathcal{D}^t$ and previous data $\mathcal{D}^{t-1}$.
\begin{align*}
s(\mathcal{D}^t, &\mathcal{D}^{t-1}; \theta^t) \\ &=
\frac{\sum_{c \in \mathcal{C}_0} \left( q_{\mathbf{x}^t}(c; \theta^t) \cdot q_{\mathbf{x}^{t-1}}(c; \theta^{t}) \right)}
{\sqrt{\sum_{c \in \mathcal{C}_0} \left( q_{\mathbf{x}^t}(c; \theta^t) \right)^2} \cdot \sqrt{\sum_{c \in \mathcal{C}_0} \left( q_{\mathbf{x}^{t-1}}(c; \theta^t) \right)^2}}
\end{align*}
The model adapts when both the uncertainty and similarity conditions are satisfied, thereby ensuring that it responds effectively.

\subsection{Pseudo-labeling}
To update the model without a ground-truth label in the post-training phase, a reliable pseudo-label must be generated. The pseudo-label \( \tilde{y}_i^t \) is assigned based on both the entropy \( h(\mathbf{x}_i^t; \theta^t) \) and confidence measure \( \Delta_{\text{MD}} \), ensuring that reliable pseudo-labels are generated only if certain conditions are satisfied.
\begin{align}
\label{eqn:pseudo-label}
\tilde{y}_i^t = 
\begin{cases} 
\hat{y}_i^t, & \text{if } h(\mathbf{x}_i^t; \theta^t) < \phi_{\text{pred}} \\[1.5ex]
\arg\min\limits_{c\in \mathcal{C}} D_{\text{MD}}(\mathbf{x}_i^t, y_i^t, \mu_c, \Sigma_c; \theta^t), 
  & \begin{array}{l}
      \text{if } h(\mathbf{x}_i^t; \theta^t) \geq \phi_{\text{pred}} \\
      \text{and } \Delta_{\text{MD}} \geq \phi_{\Delta_{\text{MD}}}
    \end{array} \\[1.5ex]
\text{No pseudo-labeling}, & \text{otherwise}
\end{cases}
\end{align}
\textbf{Model Prediction-based label ($h(\mathbf{x}_i^t; \theta^t) < \phi_{\text{pred}}$):} If the entropy \( h(\mathbf{x}_i^t; \theta^t) \) is below the threshold \( \phi_{\text{pred}} \), the model is confident in its prediction. Here, we directly use the prediction \( \hat{y}_i^t \) as the pseudo-label, i.e., $\tilde{y}_i^t = \hat{y}_i^t$. 

\noindent \textbf{Representation-based label ($h(\mathbf{x}_i^t; \theta^t) \geq \phi_{\text{pred}}$):} When model entropy \( h(\mathbf{x}_i^t; \theta^t) \) exceeds \( \phi_{\text{pred}} \), we cannot rely on the model's prediction. In this case, we leverage the feature representation at $\bar L$-th layer by measuring the distance to the centroid of all classes. 
Let $\mathcal M$ be a set of MD for a sample $(\mathbf{x}_i^t, y_i^t)$ for all classes.
\begin{align*}
\mathcal{M} = \left\{ D_{\text{MD}}(\mathbf{x}_i^t,y_i^t, \mu_c, \Sigma_c; \theta^t) | c \in \mathcal C_0 \right\}
\end{align*}
The set $\mathcal M$ contains $|\mathcal C|$ elements. 
A straightforward approach is to assign a pseudo-label to the class with the smallest distance, i.e., $\tilde y_i^t = \arg\min_{c\in \mathcal C} D_{\text{MD}}(\mathbf{x}_i^t, \mu_c, \Sigma_c; \theta^t)$.
To ensure a more reliable pseudo-label, we further assess the confidence measure $\Delta_{\text{MD}}$, which represents the distance gap between the first and second closest class centroids. Crucially, the ability to assign reliable pseudo-labels stems from the borderline refinement process in pre-training, which increases the distance between competing class centroids for ambiguous samples.
\begin{align*}
\Delta_{\text{MD}} =  \min\left(\mathcal{M} \setminus \{\min(\mathcal{M})\}\right)-\min(\mathcal{M}) 
\end{align*}
Here, $\min(\cdot)$ returns the element with the minimum value in the set and $\setminus$ denotes the set difference. We employ the smallest distance class as a pseudo-label only if the confidence measure $\Delta_{\text{MD}}$ exceeds \( \phi_{\text{MD}} \). 
If any of these conditions are not met, we avoid pseudo-labeling and exclude the sample from the post-training process.

After completing the pseudo-labeling process, we train the learnable parameters $l$ using cross-entropy loss in~\eqref{equ2}, as follows.
\begin{equation}
    l^{t} \leftarrow l^{t-1} - \eta \nabla \mathcal{L}_{\text{class}}(\mathbf{x}_i^t, \tilde{y}_i^t; \theta^{t-1})
    \label{eq:post}
\end{equation}
Here, \( \eta \) denotes the learning rate, and \( \tilde{y}_i^t \) represents the assigned pseudo-label for sample \( \mathbf{x}_i^t \) at timestep \( t \).

\subsection{Inference with Unseen Detection}
At each timestep, inference is performed after completing post-training, incorporating unseen detection as follows.
\begin{equation} 
\hat{y}_i^t = 
\begin{cases} 
c_{unseen}, & \text{if } h(\mathbf{x}_i^t; \theta^t) > \psi_{\text{pred}} \\
&\text{and } \min D_{\text{MD}} > \psi_{\text{MD}} \\
&\text{and }  \Delta_{\text{MD}} <\psi_{\Delta_{\text{MD}}}\\ 
\arg\max\limits_{c \in \mathcal{C}_0} p(c|\mathbf{x}_i^t; \theta^t), & \text{otherwise} 
\end{cases} 
\label{eq:unseen_detection}
\end{equation}
Unseen samples are expected to exhibit higher entropy due to the model has not encountered them during training. They are likely to be distant from all known class centroids, resulting in a larger minimum MD (\(\min D_{\text{MD}}\)), while maintaining similar distances across multiple classes, leading to a smaller MD difference (\(\Delta_{\text{MD}}\)). Based on these characteristics, unseen detection is performed, while samples that do not meet these conditions are classified using the model’s predictions. The proposed OASIS effectively addresses open-world problems by performing unseen detection and adaptive classification that responds to data distribution shifts in a self-supervised manner based on knowledge of representation from contrastive learning.

\section{Simulation Setups}
\begin{table*}[tb]
\centering
\caption{Accuracy comparison of post-training performance (open-world setting)}
\label{tab:performance_results}
\renewcommand{\arraystretch}{1.2} 
\setlength{\tabcolsep}{4.5pt} 
\resizebox{\textwidth}{!}{
\begin{tabular}{c|c c c c|c c c c|c c c c}
\hline
\multicolumn{13}{c}{\textbf{Accuracy [$\%$]}}\\
\hline
\multirow{2}{*}{\textbf{Method}} & \multicolumn{4}{c|}{\textbf{CIFAR-10}} & \multicolumn{4}{c|}{\textbf{CIFAR-100}} & \multicolumn{4}{c}{\textbf{Tiny-ImageNet}} \\ 
 & Lin & Squ & Sin & Ber & Lin & Squ & Sin & Ber & Lin & Squ & Sin & Ber \\ \hline\hline
\textbf{Base} & 55.8{\small\textsubscript{±3.1}} & 52.3{\small\textsubscript{±2.4}} & 54.7{\small\textsubscript{±3.5}} & 73.2{\small\textsubscript{±4.2}} & 41.5{\small\textsubscript{±2.9}} & 40.8{\small\textsubscript{±4.1}} & 42.2{\small\textsubscript{±3.7}} & 40.9{\small\textsubscript{±3.3}} & 37.4{\small\textsubscript{±3.8}} & 36.9{\small\textsubscript{±2.5}} & 38.1{\small\textsubscript{±4.4}} & 37.2{\small\textsubscript{±3.1}} \\ \hline

\textbf{UDA} & 84.1{\small\textsubscript{±2.7}} & 83.5{\small\textsubscript{±3.4}} & 82.7{\small\textsubscript{±3.1}} & 82.2{\small\textsubscript{±2.9}} & 42.6{\small\textsubscript{±3.8}} & 41.9{\small\textsubscript{±3.2}} & 41.4{\small\textsubscript{±4.5}} & 41.0{\small\textsubscript{±2.6}} & 38.4{\small\textsubscript{±3.7}} & 37.8{\small\textsubscript{±4.2}} & 37.3{\small\textsubscript{±3.9}} & 36.8{\small\textsubscript{±2.5}} \\ \hline

\textbf{OSLS} & 85.0{\small\textsubscript{±3.2}} & 84.4{\small\textsubscript{±2.9}} & 83.8{\small\textsubscript{±2.5}} & 83.2{\small\textsubscript{±3.7}} & 43.2{\small\textsubscript{±4.0}} & 42.5{\small\textsubscript{±3.3}} & 41.9{\small\textsubscript{±3.7}} & 41.3{\small\textsubscript{±2.8}} & 38.9{\small\textsubscript{±3.6}} & 38.3{\small\textsubscript{±4.4}} & 37.7{\small\textsubscript{±3.2}} & 37.1{\small\textsubscript{±2.9}} \\ \hline

\textbf{ATLAS} & 86.9{\small\textsubscript{±3.5}} & 83.2{\small\textsubscript{±2.8}} & 82.6{\small\textsubscript{±3.1}} & 82.0{\small\textsubscript{±4.2}} & 44.5{\small\textsubscript{±2.9}} & 41.2{\small\textsubscript{±4.0}} & 40.7{\small\textsubscript{±3.5}} & 40.1{\small\textsubscript{±2.7}} & 39.5{\small\textsubscript{±3.2}} & 36.7{\small\textsubscript{±3.9}} & 36.2{\small\textsubscript{±4.5}} & 35.6{\small\textsubscript{±3.6}} \\ \hline

\textbf{UNIDA} & 86.3{\small\textsubscript{±2.9}} & 85.8{\small\textsubscript{±3.5}} & 85.2{\small\textsubscript{±2.6}} & 84.7{\small\textsubscript{±3.8}} & 45.3{\small\textsubscript{±3.2}} & 44.7{\small\textsubscript{±4.1}} & 44.2{\small\textsubscript{±3.4}} & 43.7{\small\textsubscript{±2.9}} & 40.1{\small\textsubscript{±3.3}} & 39.4{\small\textsubscript{±4.3}} & 38.8{\small\textsubscript{±3.9}} & 38.1{\small\textsubscript{±2.5}} \\ \hline

\textbf{OW-SSL} & 86.2{\small\textsubscript{±3.0}} & 85.7{\small\textsubscript{±2.7}} & 85.1{\small\textsubscript{±3.5}} & 84.6{\small\textsubscript{±3.2}} & 45.1{\small\textsubscript{±4.2}} & 44.6{\small\textsubscript{±3.6}} & 44.0{\small\textsubscript{±3.9}} & 43.4{\small\textsubscript{±2.8}} & 39.8{\small\textsubscript{±3.4}} & 39.2{\small\textsubscript{±4.5}} & 38.5{\small\textsubscript{±3.2}} & 38.0{\small\textsubscript{±2.6}} \\ \hline

\textbf{Ours} & \textbf{91.8}{\small\textsubscript{±1.5}} & \textbf{91.2}{\small\textsubscript{±2.4}} & \textbf{90.9}{\small\textsubscript{±3.3}} & \textbf{90.4}{\small\textsubscript{±2.8}} & \textbf{53.1}{\small\textsubscript{±3.7}} & \textbf{52.5}{\small\textsubscript{±3.9}} & \textbf{52.0}{\small\textsubscript{±4.1}} & \textbf{51.5}{\small\textsubscript{±3.6}} & \textbf{46.7}{\small\textsubscript{±3.4}} & \textbf{46.1}{\small\textsubscript{±4.0}} & \textbf{45.6}{\small\textsubscript{±3.5}} & \textbf{45.1}{\small\textsubscript{±2.7}} \\ \hline
\end{tabular}}
\end{table*}

\begin{table}[t]
\centering
\caption{Ablation study on borderline sample refinement step (open-world setting)}
\label{tab:ablation_borderline}
\renewcommand{\arraystretch}{1.2}
\setlength{\tabcolsep}{6pt}
{
\begin{tabular}{cc|c c}
\hline
\multicolumn{1}{c}{\textbf{Dataset}} & \textbf{Shift} & \textbf{w/o} & \textbf{w/} \\
\hline\hline
\multirow{4}{*}{\textbf{CIFAR-10}} 
    & Lin & 86.3{\small\textsubscript{±1.2}} & \textbf{91.8}{\small\textsubscript{±1.5}} \\
    & Squ & 85.7{\small\textsubscript{±1.9}} & \textbf{91.2}{\small\textsubscript{±2.4}} \\
    & Sin & 85.4{\small\textsubscript{±1.7}} & \textbf{90.9}{\small\textsubscript{±3.3}} \\
    & Ber & 85.0{\small\textsubscript{±1.5}} & \textbf{90.4}{\small\textsubscript{±2.8}} \\
\hline
\multirow{4}{*}{\textbf{CIFAR-100}} 
    & Lin & 49.9{\small\textsubscript{±1.0}} & \textbf{53.1}{\small\textsubscript{±3.7}} \\
    & Squ & 49.4{\small\textsubscript{±1.0}} & \textbf{52.5}{\small\textsubscript{±3.9}} \\
    & Sin & 48.9{\small\textsubscript{±0.9}} & \textbf{52.0}{\small\textsubscript{±4.1}} \\
    & Ber & 48.4{\small\textsubscript{±1.8}} & \textbf{51.5}{\small\textsubscript{±3.6}} \\
\hline
\multirow{4}{*}{\textbf{Tiny-ImageNet}} 
    & Lin & 43.9{\small\textsubscript{±1.5}} & \textbf{46.7}{\small\textsubscript{±3.4}} \\
    & Squ & 43.3{\small\textsubscript{±1.6}} & \textbf{46.1}{\small\textsubscript{±4.0}} \\
    & Sin & 42.9{\small\textsubscript{±0.8}} & \textbf{45.6}{\small\textsubscript{±3.5}} \\
    & Ber & 42.4{\small\textsubscript{±2.0}} & \textbf{45.1}{\small\textsubscript{±2.7}} \\
\hline
\end{tabular}
}
\vspace{-0.3cm}
\end{table}

We evaluate the effectiveness of our proposed method using three widely recognized datasets in open-world study. To better reflect real-world scenarios, we employ imbalanced versions of these datasets for pre-training and apply distribution shifts over time for post-training.

\subsection{Datasets}
For pre-training, we utilize CIFAR-10-LT~\cite{CIFAR10LT}, CIFAR-100-LT~\cite{CIFAR100LT}, and Tiny-ImageNet-LT~\cite{TinyImageNetLT}, which are the long-tailed versions of CIFAR-10, CIFAR-100, and ImageNet, respectively, following the standard protocol in long-tailed recognition studies. The imbalance factor~\cite{imbalancefactor} $\rho$ is defined as $\rho = \frac{\max_c n_c}{\min_c n_c}$, where $n_c$ denotes the number of samples in class $c$. In our experiments, we set the imbalance factor to 10 for all three datasets to reflect real-world class imbalance scenarios.

For post-training, we employ CIFAR-10-C, CIFAR-100-C, and ImageNet-C~\cite{Corruption}, which introduce various corruptions to the test sets of CIFAR-10, CIFAR-100, and ImageNet, respectively. These corrupted datasets contain different corruption types, each with multiple severity levels. In our setting, we use Gaussian noise to simulate distributional shifts over time.

\subsection{Distribution Shift Dynamics}
To model real-world distribution shifts, we define two distributions $\omega_0$ and $\omega_T$. The initial distribution $\omega_0$ follows the same distribution as the pre-training phase, maintaining the pre-training distribution. The final distribution $\omega_T$ undergoes modifications where corruption is introduced, label distribution changes, and unseen classes are added. The unseen classes are included in the same proportion as the minor classes to ensure balanced representation. The distribution shift between $\omega_0$ and $\omega_T$ reflects both covariate shift and label distribution changes over time~\cite{ATLAS,FTH}. We consider four types of shifts as linear (Lin), square (Squ), sine (Sin), and Bernoulli (Ber).

\subsection{Baseline Methods}
\begin{table*}[tb]
\centering
\caption{Performance comparisons in label shift}
\label{tab:label shift}
\renewcommand{\arraystretch}{1.2} 
\setlength{\tabcolsep}{4.5pt} 
\resizebox{\textwidth}{!}{
\begin{tabular}{c|c c c c|c c c c|c c c c}
\hline
\multicolumn{13}{c}{\textbf{Accuracy [$\%$]}}\\
\hline
\multirow{2}{*}{\textbf{Method}} & \multicolumn{4}{c|}{\textbf{CIFAR-10}} & \multicolumn{4}{c|}{\textbf{CIFAR-100}} & \multicolumn{4}{c}{\textbf{Tiny-ImageNet}} \\ 
 & Lin & Squ & Sin & Ber & Lin & Squ & Sin & Ber & Lin & Squ & Sin & Ber \\ \hline\hline

\textbf{Base} & 60.5{\small\textsubscript{±2.1}} & 56.8{\small\textsubscript{±3.0}} & 58.2{\small\textsubscript{±1.9}} & 76.4{\small\textsubscript{±3.8}} & 41.2{\small\textsubscript{±3.4}} & 40.6{\small\textsubscript{±3.0}} & 42.0{\small\textsubscript{±3.8}} & 40.9{\small\textsubscript{±3.5}} & 37.5{\small\textsubscript{±2.9}} & 37.1{\small\textsubscript{±3.7}} & 38.3{\small\textsubscript{±4.1}} & 37.8{\small\textsubscript{±3.2}} \\ \hline

\textbf{UDA} & 85.2{\small\textsubscript{±2.5}} & 84.7{\small\textsubscript{±2.9}} & 84.0{\small\textsubscript{±2.0}} & 83.6{\small\textsubscript{±2.6}} & 44.1{\small\textsubscript{±3.6}} & 43.5{\small\textsubscript{±2.8}} & 43.0{\small\textsubscript{±3.5}} & 42.6{\small\textsubscript{±3.3}} & 39.2{\small\textsubscript{±3.3}} & 38.7{\small\textsubscript{±4.0}} & 38.2{\small\textsubscript{±4.2}} & 37.6{\small\textsubscript{±3.5}} \\ \hline

\textbf{OSLS} & 86.5{\small\textsubscript{±2.0}} & 85.9{\small\textsubscript{±2.8}} & 85.2{\small\textsubscript{±2.4}} & 84.7{\small\textsubscript{±2.5}} & 45.5{\small\textsubscript{±3.2}} & 44.8{\small\textsubscript{±3.0}} & 44.2{\small\textsubscript{±3.6}} & 43.6{\small\textsubscript{±3.1}} & 40.1{\small\textsubscript{±3.4}} & 39.5{\small\textsubscript{±4.1}} & 38.9{\small\textsubscript{±3.8}} & 38.3{\small\textsubscript{±3.6}} \\ \hline

\textbf{ATLAS} & 87.4{\small\textsubscript{±2.1}} & 85.8{\small\textsubscript{±3.0}} & 85.1{\small\textsubscript{±2.7}} & 84.5{\small\textsubscript{±2.9}} & 47.1{\small\textsubscript{±3.5}} & 42.8{\small\textsubscript{±3.2}} & 42.2{\small\textsubscript{±3.4}} & 41.7{\small\textsubscript{±3.6}} & 41.0{\small\textsubscript{±3.6}} & 38.3{\small\textsubscript{±4.0}} & 37.7{\small\textsubscript{±3.7}} & 37.1{\small\textsubscript{±3.4}} \\ \hline

\textbf{UNIDA} & 87.1{\small\textsubscript{±2.3}} & 86.6{\small\textsubscript{±2.8}} & 86.0{\small\textsubscript{±2.5}} & 85.5{\small\textsubscript{±2.7}} & 48.3{\small\textsubscript{±3.4}} & 47.7{\small\textsubscript{±3.2}} & 47.1{\small\textsubscript{±3.7}} & 46.5{\small\textsubscript{±3.3}} & 42.8{\small\textsubscript{±3.5}} & 42.1{\small\textsubscript{±3.9}} & 41.4{\small\textsubscript{±3.8}} & 40.8{\small\textsubscript{±3.6}} \\ \hline

\textbf{OW-SSL} & 86.9{\small\textsubscript{±2.2}} & 86.4{\small\textsubscript{±2.9}} & 85.8{\small\textsubscript{±2.6}} & 85.3{\small\textsubscript{±2.8}} & 48.0{\small\textsubscript{±3.6}} & 47.4{\small\textsubscript{±3.1}} & 46.8{\small\textsubscript{±3.6}} & 46.2{\small\textsubscript{±3.2}} & 42.5{\small\textsubscript{±3.7}} & 41.9{\small\textsubscript{±3.8}} & 41.2{\small\textsubscript{±4.0}} & 40.6{\small\textsubscript{±3.5}} \\ \hline

\textbf{Ours} & \textbf{92.1}{\small\textsubscript{±1.8}} & \textbf{91.6}{\small\textsubscript{±2.6}} & \textbf{91.2}{\small\textsubscript{±2.3}} & \textbf{90.8}{\small\textsubscript{±2.0}} & \textbf{55.3}{\small\textsubscript{±3.0}} & \textbf{54.7}{\small\textsubscript{±3.5}} & \textbf{54.1}{\small\textsubscript{±3.2}} & \textbf{53.6}{\small\textsubscript{±3.3}} & \textbf{49.3}{\small\textsubscript{±3.6}} & \textbf{48.7}{\small\textsubscript{±3.9}} & \textbf{48.2}{\small\textsubscript{±3.8}} & \textbf{47.6}{\small\textsubscript{±3.4}} \\ \hline

\end{tabular}}
\end{table*}

\begin{table*}[tb]
\centering
\caption{Performance comparisons in covariate shift}
\label{tab:domain shift}
\renewcommand{\arraystretch}{1.2} 
\setlength{\tabcolsep}{4.5pt} 
\resizebox{\textwidth}{!}{
\begin{tabular}{c|c c c c|c c c c|c c c c}
\hline
\multicolumn{13}{c}{\textbf{Accuracy [$\%$]}}\\
\hline
\multirow{2}{*}{\textbf{Method}} & \multicolumn{4}{c|}{\textbf{CIFAR-10}} & \multicolumn{4}{c|}{\textbf{CIFAR-100}} & \multicolumn{4}{c}{\textbf{Tiny-ImageNet}} \\ 
 & Lin & Squ & Sin & Ber & Lin & Squ & Sin & Ber & Lin & Squ & Sin & Ber \\ \hline\hline

\textbf{Base} & 59.5{\small\textsubscript{±2.3}} & 56.0{\small\textsubscript{±3.2}} & 57.8{\small\textsubscript{±2.0}} & 76.3{\small\textsubscript{±4.0}} & 43.8{\small\textsubscript{±3.3}} & 43.1{\small\textsubscript{±2.9}} & 44.5{\small\textsubscript{±4.0}} & 43.3{\small\textsubscript{±3.7}} & 39.2{\small\textsubscript{±2.7}} & 38.8{\small\textsubscript{±3.5}} & 40.1{\small\textsubscript{±4.2}} & 39.3{\small\textsubscript{±3.0}} \\ \hline

\textbf{UDA} & 85.5{\small\textsubscript{±2.3}} & 84.9{\small\textsubscript{±3.1}} & 84.2{\small\textsubscript{±1.9}} & 83.7{\small\textsubscript{±2.8}} & 44.6{\small\textsubscript{±3.9}} & 43.9{\small\textsubscript{±2.5}} & 43.4{\small\textsubscript{±3.8}} & 43.0{\small\textsubscript{±3.2}} & 39.8{\small\textsubscript{±3.4}} & 39.2{\small\textsubscript{±4.0}} & 38.6{\small\textsubscript{±4.3}} & 38.1{\small\textsubscript{±3.7}} \\ \hline

\textbf{OSLS} & 86.3{\small\textsubscript{±1.9}} & 85.8{\small\textsubscript{±2.8}} & 85.2{\small\textsubscript{±2.3}} & 84.6{\small\textsubscript{±2.6}} & 45.2{\small\textsubscript{±3.1}} & 44.5{\small\textsubscript{±2.9}} & 43.9{\small\textsubscript{±3.5}} & 43.3{\small\textsubscript{±3.0}} & 40.4{\small\textsubscript{±3.5}} & 39.7{\small\textsubscript{±4.2}} & 39.1{\small\textsubscript{±3.9}} & 38.5{\small\textsubscript{±3.6}} \\ \hline

\textbf{ATLAS} & 88.2{\small\textsubscript{±2.0}} & 84.5{\small\textsubscript{±3.3}} & 83.9{\small\textsubscript{±2.5}} & 83.3{\small\textsubscript{±2.7}} & 46.8{\small\textsubscript{±3.7}} & 42.4{\small\textsubscript{±3.1}} & 41.9{\small\textsubscript{±3.2}} & 41.3{\small\textsubscript{±3.4}} & 41.7{\small\textsubscript{±3.9}} & 38.8{\small\textsubscript{±4.1}} & 38.2{\small\textsubscript{±3.7}} & 37.5{\small\textsubscript{±3.3}} \\ \hline

\textbf{UNIDA} & 87.8{\small\textsubscript{±2.2}} & 87.2{\small\textsubscript{±3.0}} & 86.6{\small\textsubscript{±2.6}} & 86.1{\small\textsubscript{±2.4}} & 47.3{\small\textsubscript{±3.2}} & 46.7{\small\textsubscript{±3.1}} & 46.2{\small\textsubscript{±3.8}} & 45.7{\small\textsubscript{±3.4}} & 42.6{\small\textsubscript{±3.4}} & 41.9{\small\textsubscript{±4.0}} & 41.3{\small\textsubscript{±3.9}} & 40.6{\small\textsubscript{±3.6}} \\ \hline

\textbf{OW-SSL} & 87.7{\small\textsubscript{±2.1}} & 87.2{\small\textsubscript{±2.9}} & 86.6{\small\textsubscript{±2.5}} & 86.1{\small\textsubscript{±2.7}} & 47.1{\small\textsubscript{±3.5}} & 46.6{\small\textsubscript{±3.0}} & 46.0{\small\textsubscript{±3.7}} & 45.4{\small\textsubscript{±3.3}} & 42.3{\small\textsubscript{±3.8}} & 41.7{\small\textsubscript{±3.9}} & 41.0{\small\textsubscript{±4.1}} & 40.5{\small\textsubscript{±3.5}} \\ \hline

\textbf{Ours} & \textbf{92.5}{\small\textsubscript{±1.9}} & \textbf{92.0}{\small\textsubscript{±2.7}} & \textbf{91.7}{\small\textsubscript{±2.4}} & \textbf{91.2}{\small\textsubscript{±2.1}} & \textbf{54.3}{\small\textsubscript{±3.1}} & \textbf{53.7}{\small\textsubscript{±3.6}} & \textbf{53.2}{\small\textsubscript{±3.3}} & \textbf{52.7}{\small\textsubscript{±3.4}} & \textbf{48.8}{\small\textsubscript{±3.7}} & \textbf{48.2}{\small\textsubscript{±4.0}} & \textbf{47.7}{\small\textsubscript{±3.9}} & \textbf{47.1}{\small\textsubscript{±3.5}} \\ \hline

\end{tabular}}
\end{table*}

We evaluate our proposed method by comparing it with existing approaches that address different aspects of distribution shifts, as follows.
\begin{itemize}
    \item \textbf{Base}: A method that trains the model solely with the pre-training phase, serving as a baseline without any post-training.
    \item \textbf{UDA}~\cite{UDA}: A widely used method for unsupervised domain adaptation that addresses covariate shift, but does not consider seen class detection, class imbalance, or label shift.
    
    \item \textbf{ATLAS}~\cite{ATLAS}: A label shift-aware method that neglects covariate shift, class imbalance, and seen/unseen class separation.

    \item \textbf{OSLS}~\cite{OSLS}: A method designed for seen class detection with label shift, yet it does not account for covariate shift or class imbalance.
    
    \item \textbf{UNIDA}~\cite{UNIDA}: An approach that incorporates seen class detection into domain adaptation, but does not consider label shift and class imbalance.
    
    \item \textbf{OW-SSL}~\cite{OWSSL}: A semi-supervised method that focuses on seen class detection with covariate shift and label shift, but does not handle class imbalance.
    
    \item \textbf{Ours}: A proposed framework that jointly addresses covariate shift, label shift, unseen class detection, and class imbalance for robust open-world adaptation.
\end{itemize}
For a fair comparison, we utilize the same pre-trained models for all baseline methods. Our method distinguishes itself by jointly handling covariate shift, seen detection, class imbalance, and label shift, making it a more comprehensive approach for real-world scenarios.

\subsection{Implementation Details}

We utilize ResNet-18 as the backbone network for all three datasets CIFAR-10, CIFAR-100, and ImageNet. The same architecture is applied across all experiments for consistency. The pre-trained models are obtained using our proposed training method, and all baseline comparison models are evaluated using the same pre-trained models.

\section{Simulation Results}
\noindent{\bf{Open-world Setting}}: 
As shown in Table~\ref{tab:performance_results}, we evaluate the methods in an open-world setting where label shift, covariate shift, and unseen class emergence occur simultaneously. Most existing methods struggle to adapt in this challenging scenario, as they are designed to handle only one or two aspects of the open-world problem. Our proposed method achieves significantly higher performance by jointly tackling label shift, covariate shift, and unseen class detection. Across all datasets and shift scenarios, our method achieves an average relative improvement rate of 13.74\% over the best-performing existing method. This confirms that our approach provides a more comprehensive solution for real-world distribution shifts.

We also conduct an ablation study on the borderline sample refinement step to investigate its impact on post-training performance under open-world conditions. As shown in Table~\ref{tab:ablation_borderline}, enabling borderline refinement consistently improves performance across all datasets and shift types, yielding an average relative improvement rate of 6.2\%. These results demonstrate that refining borderline samples during pre-training leads to more reliable representations and decision boundaries, supporting more effective pseudo-labeling in the post-training stage. This validates our hypothesis that better initial representations—especially around class boundaries—are critical for robust adaptation in open-world settings.

\noindent{\bf{Label Shift}}: 
As shown in Table~\ref{tab:label shift}, we compare the performances of different methods under label shift scenarios. The results indicate that existing baselines struggle to adapt to significant changes in label distributions with class imbalance, particularly in cases where rare classes become more dominant. Our approach achieves superior performance across various label distribution shifts by effectively adjusting to changing label distributions and incorporating unseen classes, achieving an average relative improvement of 12.18\% over the best-performing baseline in each scenario. These results highlight the robustness of our imbalance-aware contrastive learning step for adapting to dynamic label distribution shifts under class imbalance.

\noindent{\bf{Covariate Shift}}: 
Table~\ref{tab:domain shift} presents the comparison results under covariate shift scenarios. Our method achieves the highest performance across all datasets and scenarios, with an average relative improvement rate of 11.92\% over the next best-performing method. This significant gain demonstrates the effectiveness of our approach in handling covariate shift under class-imbalanced conditions, where existing methods often struggle due to insufficient representation learning within the same domain. By constructing a more discriminative and balanced representation space during pre-training, our method enables better generalization in the post-training phase.

\section{Conclusions}
In this work, we introduced a comprehensive framework for handling complex open-world scenarios where label shift, covariate shift, and the emergence of unseen classes occur simultaneously. For the pre-training phase, we propose a straightforward yet effective refinement step, which enhances the decision boundary for each class within the representation space, establishing a solid foundation for better adaptation during post-training. Our post-training method enabled adaptive model updates under dynamic conditions. Through extensive experiments across multiple datasets, we demonstrated that our approach consistently outperforms existing state-of-the-art methods in handling diverse distributional shifts. Furthermore, the observed performance gap in the ablation study demonstrates that the borderline refinement step is essential for generating a reliable pseudo-label. The results validate the effectiveness of our framework in real-world applications, where adaptability to changing data distributions is crucial.

\appendix
\section*{Appendix}

\begin{table*}[ht]
    \caption{Notation Table}
    \label{tab:notation}
    \centering
    \begin{tabular}{llll}
    \hline
    \textbf{Notation} & \textbf{Description} & \textbf{Notation} & \textbf{Description} \\ \hline
    $\mathcal{D}_0$ & Labeled pre-training dataset & $\mathcal{D}_t$ & Unlabeled post-training dataset at time $t$ \\
    $\mathbf{x}_i^0$ & $i$-th input sample in pre-training & $y_i^0$ & Label for $\mathbf{x}_i^0$ \\
    $\mathbf{x}_i^t$ & $i$-th input sample at time $t$ & $N_0$ & Number of pre-training samples \\
    $N_t$ & Number of post-training samples at $t$ & $\rho$ & Class imbalance factor \\
    $\mathcal{C}_0$ & Set of known classes & $\mathcal{C}_{all}$ & Set of all possible classes \\
    $c$ & Class index & $|\mathcal{C}_0|$ & Number of known classes \\
    $\omega_0(c)$ & Initial label distribution & $\omega_T(c)$ & Final label distribution \\
    $\Omega^t(c)$ & Label distribution at time $t$ & $\alpha^t$ & Time-dependent parameter \\
    $T$ & Total number of timesteps & $t$ & Current timestep \\
    $\theta^0$ & Pre-trained model parameters & $\theta^t$ & Model parameters at time $t$ \\
    $f$ & Frozen parameters & $l$ & Learnable parameters \\
    $\theta_{:\bar{L}}$ & Model parameters up to $\bar{L}$-th layer & $L$ & Total number of layers \\
    $\bar{L}$ & Representation layer & $\mathcal{L}_{\text{pre}}$ & Pre-training loss \\
    $\mathcal{L}_{\text{post}}$ & Post-training loss & $\mathcal{L}_{\text{class}}$ & Cross-entropy loss \\
    $\mathcal{L}_{\text{rep}}$ & Representation loss & $\lambda$ & Loss scaling factor \\
    $\epsilon$ & Margin in representation loss & $p_{\mathbf{x}_i^0}(c)$ & True class probability \\
    $q_{\mathbf{x}_i^t}(c;\theta^t)$ & Predicted probability & $\eta$ & Learning rate \\
    $\mu_c$ & Mean vector for class $c$ & $\Sigma_c$ & Covariance matrix for class $c$ \\
    $D_{\text{MD}}(\cdot)$ & Mahalanobis distance & $\Delta_{\text{MD}}$ & MD difference \\
    $\phi_{\text{border}}$ & Borderline threshold & $\phi_{\text{MD}}$ & MD threshold \\
    $\phi_{\text{ent}}$ & Entropy threshold & $\phi_{\text{cos}}$ & Cosine similarity threshold \\
    $\phi_{\text{pred}}$ & Pseudo-label confidence threshold & $\psi_{\text{pred}}$ & Entropy threshold \\
    $\psi_{\text{MD}}$ & Inference MD threshold & $\psi_{\Delta_{\text{MD}}}$ & Inference MD-difference threshold \\
    $\mathbf{x}_{\text{a},c}^0$ & Anchor sample for class $c$ & $\mathbf{x}_{\text{b},c}^0$ & Borderline sample for class $c$ \\
    $y_{\text{a},c}^0$ & Anchor label for class $c$ & $y_{\text{b},c}^0$ & Borderline label for class $c$ \\
    $h(\mathbf{x}_i^t; \theta^t)$ & Entropy of prediction for $\mathbf{x}_i^t$ & $s(\mathcal{D}_t, \mathcal{D}_{t-1}; \theta^t)$ & Cosine similarity between distributions \\
    $\tilde{y}_i^t$ & Pseudo-label for $\mathbf{x}_i^t$ & $\hat{y}_i^t$ & Model predicted label for $\mathbf{x}_i^t$ \\
    $c_{unseen}$ & Unseen class label & $\mathcal{M}$ & Set of MDs for all classes \\
    \hline
    \end{tabular}
\end{table*}

\section{Simulation Settings}
\label{sec:simul_setup}

This section describes the experimental setup used to evaluate the performance of our method under different distribution shifts. We first define the types of shifts we simulate, followed by the model architectures and hardware specifications used in our experiments.

\begin{figure}[h]
    \centering
    \includegraphics[width=\columnwidth]{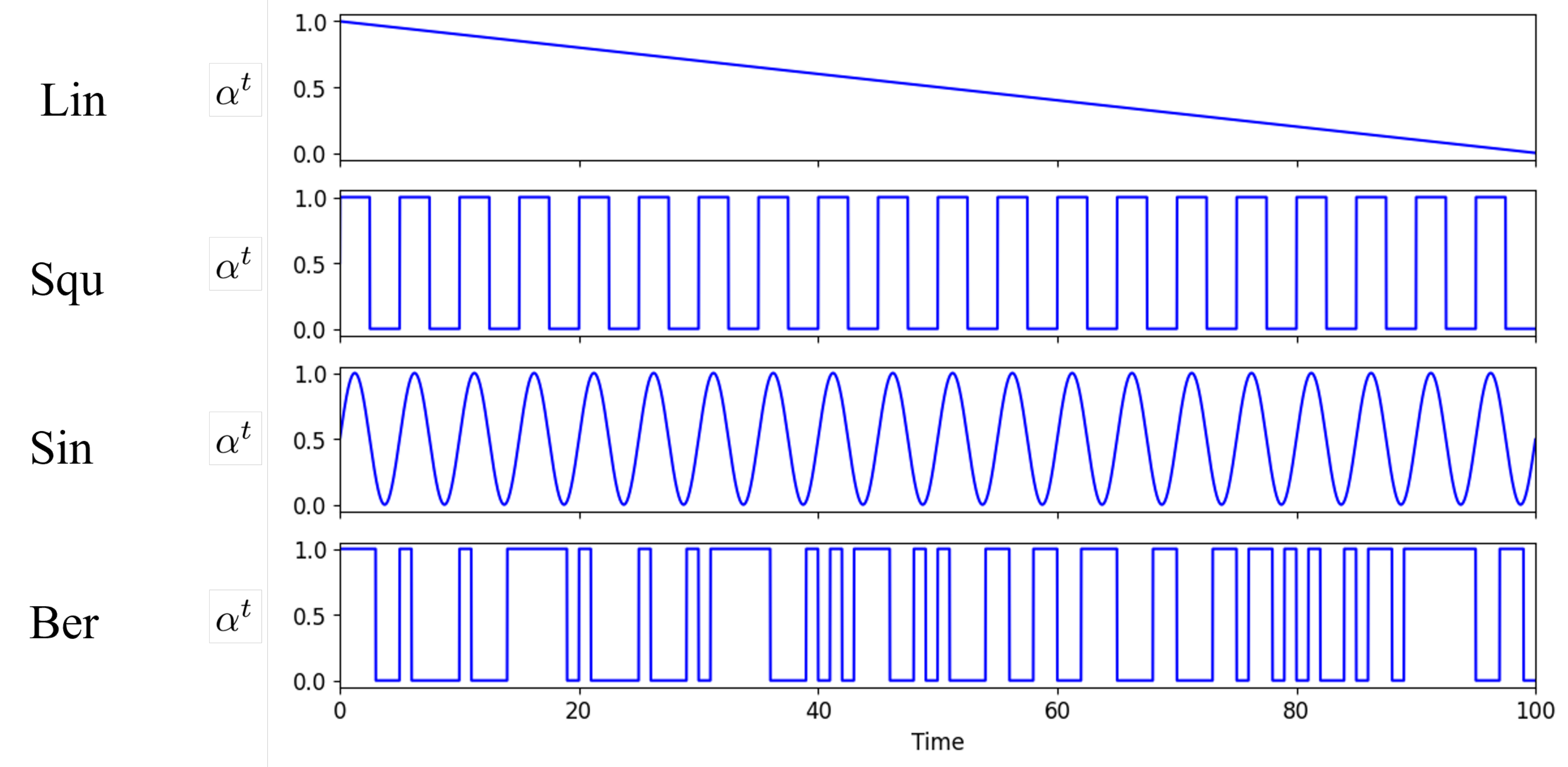}
    \caption{Illustration of how \(\alpha^t\) changes over time under different shift types. Square shift and sine shift exhibit periodic behaviors, while linear shift maintains a smooth transition. The Bernoulli shift demonstrates a stochastic behavior, making it less predictable.}
    \label{fig:shift}
\end{figure}

\begin{table*}[t]
    \centering
    \caption{Common Hyperparameter settings for simulation datasets.}
    \label{tab:com_hyperparameters}
    \renewcommand{\arraystretch}{1.2}
    \setlength{\tabcolsep}{10pt}
    \begin{tabular}{lc|lc|lc}
    \hline
    \textbf{Hyperparameter} & \textbf{Value} & \textbf{Hyperparameter} & \textbf{Value} & \textbf{Hyperparameter} & \textbf{Value} \\
    \hline
        Learning Rate & 0.0001 & Batch Size & 128 & Pre-train Epochs & 100 \\
        Model & ResNet-18 & Weight Balance $\lambda$ & 0.25 & Post-training Epochs & 10 \\
        Margin $\epsilon$ & 10.0 & Threshold $\phi_{\text{MD}}$ & 3.0 & Time $T$ & 100 \\
    \hline
    \end{tabular}
\end{table*}


\begin{table*}[t]
    \centering
    \caption{Hyperparameter settings for simulation datasets.}
    \label{tab:hyperparameters}
    \renewcommand{\arraystretch}{1.2}
    \setlength{\tabcolsep}{10pt}
    \begin{tabular}{c|lc|lc|lc}
    \hline
    \textbf{Dataset} & \textbf{Hyperparameter} & \textbf{Value} & \textbf{Hyperparameter} & \textbf{Value} & \textbf{Hyperparameter} & \textbf{Value} \\
    \hline
    {CIFAR-10}
        & Threshold $\phi_{\text{pred}}$ & 0.1 & Threshold $\phi_{\text{cos}}$ & 0.5 & Threshold $\phi_{\text{ent}}$ & 0.5 \\
    \hline
    {CIFAR-100}
        & Threshold $\phi_{\text{pred}}$ & 0.3 & Threshold $\phi_{\text{cos}}$ & 0.4 & Threshold $\phi_{\text{ent}}$ & 0.4 \\
    \hline
    {Tiny-ImageNet}
        & Threshold $\phi_{\text{pred}}$ & 0.2 & Threshold $\phi_{\text{cos}}$ & 0.6 & Threshold $\phi_{\text{ent}}$ & 0.7 \\
    \hline
    \end{tabular}
\end{table*}

\subsection{Distribution Shifts}
To simulate real-world distribution shifts, we introduce a time-dependent parameter \(\alpha^t\), which governs the transition from an initial distribution \(\omega_0\) to a target distribution \(\omega_T\). We explore four types of shifts. Each of these mechanisms introduces different dynamics in how the distributions evolve over time.

\begin{itemize}
    \item \textbf{Linear Shift (Lin)}: A smooth and gradual transition where \(\alpha^t\) increases linearly over time, defined as \(\alpha^t = \frac{t}{T}\). This shift models scenarios with incremental changes, such as seasonal variations in data distribution.
    
    \item \textbf{Square Shift (Squ)}: A step-like transition where \(\alpha^t\) alternates between 0 and 1 at intervals of \(\sqrt{T}/2\). This results in abrupt, periodic distribution changes, mimicking scheduled updates or policy changes.
    
    \item \textbf{Sine Shift (Sin)}: A periodic transition given by \(\alpha^t = \sin\left(\frac{t\pi}{\sqrt{T}}\right)\), simulating cyclical variations such as daily or weekly trends in streaming data.
    
    \item \textbf{Bernoulli Shift (Ber)}: A stochastic shift where \(\alpha^t\) retains its previous value \(\alpha^{t-1}\) with probability \(\frac{1}{\sqrt{T}}\) or flips to \(1 - \alpha^{t-1}\). This shift models unpredictable distribution changes, often seen in adversarial or rapidly evolving environments.
\end{itemize}

Figure~\ref{fig:shift} illustrates the behavior of these shift mechanisms. The linear shift maintains a gradual and smooth transition, while the square shift and sine shift exhibit structured periodic patterns. The Bernoulli shift introduces stochastic variations, making it unpredictable.

\subsection{Hyperparameter Settings}
\label{sec:model_architecture}

To ensure fair comparisons and effective model adaptation, we define a set of hyperparameters tailored to each dataset. The primary goal of these settings is to balance training stability and adaptability to dynamic shifts in data distribution. For all datasets, we utilize the Adam optimizer with a fixed learning rate and apply a weight balancing factor to enhance model generalization. 

Given the varying complexity of datasets, we adjust key hyperparameters accordingly. The learning rate, batch size, model, and training epochs remain consistent across datasets, whereas threshold values and hidden layer configurations are fine-tuned to better accommodate dataset-specific characteristics. Detailed hyperparameter configurations for CIFAR-10, CIFAR-100, and Tiny-ImageNet are provided in Table~\ref{tab:com_hyperparameters} and~\ref{tab:hyperparameters}.



\newpage
\section*{Acknowledgment}
This research was supported in part by National Research Foundation of Korea (NRF) grant (RS-2023-00278812, RS-2025-02214082), and in part by the Institute of Information \& communications Technology Planning \& Evaluation (IITP) grants (IITP-2025-RS-2020-II201602) funded by the Korea government (MSIT).

\section*{GenAI Usage Disclosure}

We acknowledge the use of Generative AI (GenAI) tools in the preparation of this paper as follows:

\begin{itemize}
    \item \textbf{Writing assistance:} ChatGPT (OpenAI) was used for improving grammar, rephrasing, and refining the clarity of certain paragraphs. All substantive content and structure were authored by the authors.
    \item \textbf{Code generation:} No GenAI tools were used to generate or write code used in this study.
    \item \textbf{Data processing or analysis:} No GenAI tools were used for data processing, analysis, or result generation.
\end{itemize}

All uses of GenAI tools complied with the ACM Authorship Policy on Generative AI usage.

\bibliographystyle{ACM-Reference-Format}
\bibliography{ref}

\end{document}